\documentclass[runningheads]{llncs}
\usepackage{graphicx}
\begin{document}
\title{Deep-CLASS at ISIC Machine Learning Challenge 2018}
%
%
\author{Sara Nasiri,
Matthias Jung,
Julien Helsper,
Madjid Fathi}
\authorrunning{S. Nasiri et al.}
%
\institute{Department of Electrical Engineering \& Computer Science, University of Siegen   \\                                                       
	H\"olderlinstr. 3, 57076 Siegen, Germany
	\\
	\email{sara.nasiri@uni-siegen.de, (julien.helsper,matthias.jung)@student.uni-siegen.de, madjid.fathi@uni-siegen.de
}}

\maketitle              
\begin{abstract}
This paper reports the method and evaluation results of MedAusbild team for ISIC challenge task. Since early 2017, our team has worked on melanoma classification \cite{MedAusbild} \cite{Helsper2017}, and has employed deep learning since beginning of 2018 \cite{Nasiri2018}. Deep learning helps researchers absolutely to treat and detect diseases by analyzing medical data (e.g., medical images). One of the representative models among the various deep-learning models is a convolutional neural network (CNN). Although our team has an experience with segmentation and classification of benign and malignant skin-lesions, we have participated in the task 3 of ISIC Challenge 2018 for classification of seven skin diseases, explained in this paper. 

\end{abstract}
\section{Introduction and Dataset}
Various skin lesions classification systems have been developed using support vector machines (SVMs) and k nearest neighbors (k-NNs) and recently deep learning (DL).
The goal of this recurring challenge is "to help participants develop image analysis tools to enable the automated diagnosis of melanoma from dermoscopic images" \cite{ISIC}. 

The training and evaluation dermoscopic images are from the ISIC Machine Learning Challenges 2018 (ISIC 2018: Skin Lesion Analysis Towards Melanoma Detection). 
There are 8010 samples in the training dataset for seven disease categories which are Melanoma (M), Melanocytic nevus (N), Basal cell carcinoma (BCC), Actinic keratosis / Bowen’s disease- intraepithelial carcinoma (AK), Benign keratosis- solar lentigo / seborrheic keratosis / lichen planus-like keratosis (PBK), Dermatofibroma (D) and Vascular lesion (VL) \cite{Task3}. For the evaluation phase, 161 samples were selected from the training dataset. The lesion images come from the HAM10000 Dataset. Data within this archive can be accessed through the archive gallery as well as through standardized API-calls \footnote{https://isic-archive.com/api/v1} and during the ISIC 2018 challenge image data, diagnosis, and the type of ground-truth was available for download.
This challenge is broken into three separate tasks in this year’s competition:
\begin{itemize}
	\item[-] Task 1: Lesion Segmentation,  
	\item[-] Task 2: Lesion Attribute Detection, and
	\item[-] Task 3: Disease Classification.
\end{itemize}

\begin{figure} [h!]
	\centering
	\includegraphics[width=0.9\textwidth]{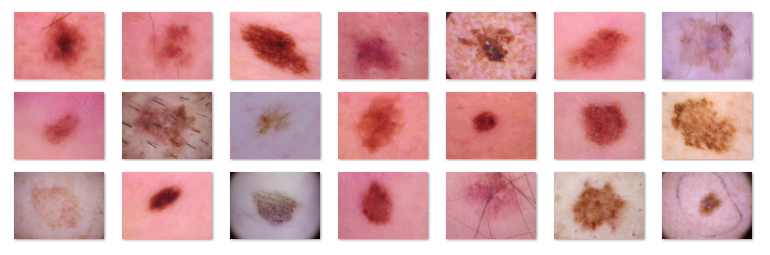}
	\caption{Examples of images for these seven diseases.} \label{fig0}
\end{figure}

In this work, we focus on Task 3. Our data was extracted from the “ISIC 2018: Skin Lesion Analysis Towards Melanoma Detection” grand challenge datasets \cite{Noel} \cite{Tschandl}.
During training, the samples (see in Fig. \ref{fig0}) are augmented by rotation and 	flipping. The amount of augmented images for these diseases are as follows:
M:13350; N:26820; BCC:16440; AK:13050; PBK:13185; D:10212; VL:11300. 
\section{Methodology and Proposed System}
Convolutional neural networks (CNNs), are a specialized kind of neural network for processing data which employs a mathematical operation called "convolution" \cite{CNN}. 

In this study, we applied an end-to-end CNN framework (machine learning system) to detect malignant lesions using images from ISIC archive dataset. We applied different types of optimization and selected the best combination of fine-tuned CNNs.
As shown in Fig.~\ref{fig1} (see Appendix), we built a 19-layer model which contains eleven convolutional (conv), five max-pooling and three fully connected (FC) layers. The input image is the first layer ($h\times w\times d$ which $h\times w$ is the pixel size and $d$ is the color channel, here is $128\times 128\times 3$). 
For applying deep-learning, we have utilized Caffe\footnote{http://caffe.berkeleyvision.org/}\cite{Jia2014} which is a deep learning framework developed by Berkeley AI Research (BAIR)\footnote{http://bair.berkeley.edu/}.
Matlab (2017a) was utilized to develop Deep-CLASS, in particular with the use of Image Processing Toolbox, Parallel Computing Toolbox, Matlab Compiler and Coder, and App Designer. 

\section{Evaluation and Results}
\begin{figure} [h!]
	\centering
	\includegraphics[width=0.9\textwidth]{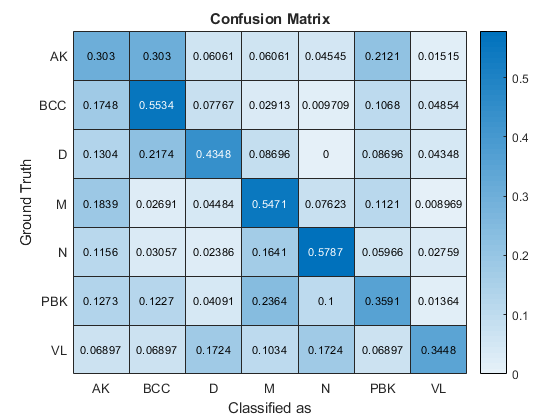}
	\caption{Confusion matrix of the unbalanced results.} \label{fig2}
\end{figure}

The performance of Deep-CLASS in terms of comparison of the evaluation scores in this task \cite{Task3} is shown in Table \ref{table:Comp_Evaluation-Melanoma}. The confusion matrix is also illustrated in Fig. \ref{fig2}.
\begin{table} [h!]
	\centering
	\caption {The comparison of evaluation scores (precision, recall (sensitivity), specificity, f-measure and accuracy) for the unbalanced dataset of DeeP-CLASS.}
	\label{table:Comp_Evaluation-Melanoma}
	\begin{tabular}{l c c c c c c c c c}
		Disease Classification 
		& \quad	TP	&  FP &  TN  &   FN  & Acc. &   F-meas. & Pre. &  Rec. (Sen.) &  Spe. \\
		\hline	
		AK  & \quad 11   &  89  & 1850  &  55  &  93  &  13  &  11  &  17  &    95 \\
		BCC & \quad 54   &  110 & 1792  &  49  &  92  &  40  &  33  &  52  &    94 \\
		D   &   \quad 2  &  27  & 1955  &  21  & 98   &   8  &  7   &   9  &    99 \\
		M   & \quad 105  &  240 & 1542  & 118  &  82  &  37  &  30  &  47  &    87 \\
		N   & \quad 930  &  101 & 563   & 411  & 74   &  78  &  90  &  69  &    85 \\
		PBK & \quad 94   &  214 & 1571  & 126  & 83   &  36  &  31  &  43  &    88 \\
		VL  &  \quad 4   &  24  & 1952  &  25  & 98   &  14  &  14  &  14  &    99 \\
		\hline
	\end{tabular}
\end{table}

\section{Conclusion}
In this paper, a training and testing method were deployed for composing a comparatively accurate CNN model from sample images of ISIC archive dataset. Although further data analysis is necessary to improve its accuracy, CNN would be helpful for the classification of diseases and in particular for the early detection of skin cancers. Analysis of the results obtained by testing an ISIC dataset in task three suggests that using our case-based system for representation of cases via CNN is fit for the purpose of supporting users by providing relevant information \cite{Nasiri2018} related to each disease. Further work will involve extending the training phase by using more images and normalization methods to improve the performance of our system and increase the accuracy.

%
%
%

\begin{thebibliography}{8}
\bibitem{Helsper2017}
Helsper, J. and Jung, M. Projektgruppe "Wissensbasiertes System zur Unterstützung der medizinischen Ausbildung (MedAusbild)"- Sommersemester2017. Technical report, Institute of KBS \& KM, University of Siegen (2017).
	
\bibitem{CNN}
Goodfellow, I., Bengio, Y., Courville, A., Deep Learning, MIT Press, \url{http://www.deeplearningbook.org}, (2016).
	
\bibitem{ISIC}
ISIC Chalenge 2018. \url{https://challenge2018.isic-archive.com/} Last access at 19 Jul. 2018.

\bibitem{Task3}
ISIC Chalenge 2018- Task 3. \url{https://challenge2018.isic-archive.com/task3/} Last access at 19 Jul. 2018.

\bibitem{Jia2014}
Jia, Y., Shelhamer, E., Donahue, J., Karayev, S., Long, J., Girshick, R., Guadarrama, S., Darrell, T. Caffe: Convolutional Architecture for Fast Feature Embedding. arXiv preprint arXiv:1408.5093, (2014).

\bibitem{MedAusbild}
MedAusbild (2018). \url{https://www.eti.uni-siegen.de/ws/projekte/medausbild/index.\\html.en?lang=en} Last access at 19 Jul. 2018.

\bibitem{Nasiri2018} 
Nasiri, S., Helsper, J., Jung, M., Fathi, M., Enriching a CBR recommender system by classification of skin lesions using deep neural networks,
In proceedings of the workshop on Deep Learning at the 26th International Conference on Case-Based Reasoning, Stockholm, Sweden, 09-12 July, 86--90, (2018). 

\bibitem{Noel}
Noel C. F. Codella, David Gutman, M. Emre Celebi, Brian Helba, Michael A. Marchetti, Stephen W. Dusza, Aadi Kalloo, Konstantinos Liopyris, Nabin Mishra, Harald Kittler, Allan Halpern: “Skin Lesion Analysis Toward Melanoma Detection: A Challenge at the 2017 International Symposium on Biomedical Imaging (ISBI), Hosted by the International Skin Imaging Collaboration (ISIC)”, (2017); arXiv:1710.05006.

\bibitem{Tschandl}
Tschandl, P., Rosendahl, C. \& Kittler, H. The HAM10000 dataset, a large collection of multi-source dermatoscopic images of common pigmented skin lesions. Sci. Data 5, 180161 doi:10.1038/sdata.2018.161 (2018).

\end{thebibliography}
%

\begin{figure}
	\includegraphics[width=1.05\textwidth]{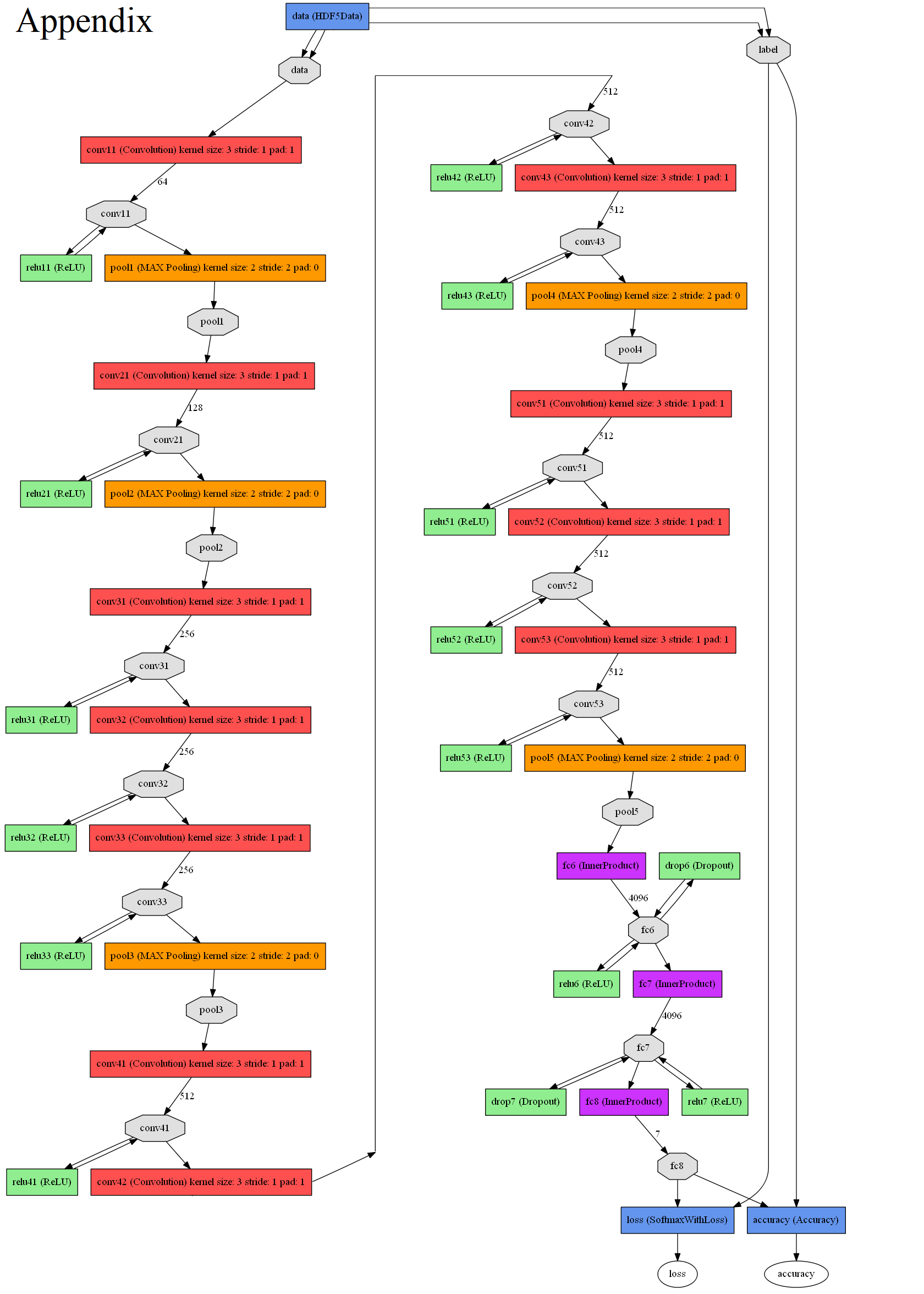}
	\caption{Layer view and a MNIST digit classification example of a Caffe network - fourteen layers excluding max-pooling; convolution in red, max-pooling in orange and fully connected layers in violet.} \label{fig1}
\end{figure}
\end{document}